\documentclass[runningheads]{llncs}
\usepackage[T1]{fontenc}
\usepackage{graphicx}
\usepackage{amsmath,amssymb} % define this before the line numbering.
\usepackage{color}
\usepackage{float}
\usepackage{tikz}
\usepackage{adjustbox}
\usepackage{wrapfig,lipsum,booktabs}
\usepackage{soul}
\usepackage{multirow}
\usepackage{subfigure}

\begin{document}
\title{Spotlights: Probing Shapes from\\Spherical Viewpoints}
\author{Jiaxin Wei\inst{1}\orcidID{0000-0002-7645-6093} \and Lige Liu\inst{2} \and Ran Cheng\inst{2} \and Wenqing Jiang\inst{1}\orcidID{0000-0002-4514-0424} \and Minghao Xu\inst{2} \and Xinyu Jiang\inst{2} \and Tao Sun\inst{2}\thanks{Corresponding author.} \and S\"oren Schwertfeger\inst{1}\orcidID{0000-0003-2879-1636} \and Laurent Kneip\inst{1}\orcidID{0000-0001-6727-6608}}
\authorrunning{J. Wei et al.}
\institute{ShanghaiTech University
\email{\{weijx,jiangwq,soerensch,lkneip\}@shanghaitech.edu.cn} \and 
RoboZone, Midea Inc\\ \email{\{liulg12,chengran1,xumh33,jxy77,tsun\}@midea.com}} 
\maketitle              % typeset the header of the contribution
\begin{abstract}
Recent years have witnessed the surge of learned representations that directly build upon point clouds. Inspired by spherical multi-view scanners, we propose a novel sampling model called \textit{Spotlights} to represent a 3D shape as a compact 1D array of depth values. It simulates the configuration of cameras evenly distributed on a sphere, where each virtual camera casts light rays from its principal point to probe for possible intersections with the object surrounded by the sphere. The structured point cloud is hence given implicitly as a function of depths. We provide a detailed geometric analysis of this new sampling scheme and prove its effectiveness in the context of the point cloud completion task. Experimental results on both synthetic and real dataset demonstrate that our method achieves competitive accuracy and consistency while at a lower computational cost. The code and dataset will be released at https://github.com/goldoak/Spotlights.
%Furthermore, we showcase one possible usage of the ordered representation on the downstream point cloud registration task.

\end{abstract}
\section{Introduction}
Over the past decade, the community has put major efforts into the recovery of explicit 3D object models, which is of eminent importance in many computer vision and robotics applications \cite{hartley03,campos2021orb,furukawa10,newcombe11,kinectfusion,shan20,fusionpp,nodeslam}. Traditionally, the recovery of complete object geometries from sensor readings relies on scanning systems that employ spherical high-resolution camera arrangements~\cite{3dscanner,multicamera}. If additionally equipped with light sources, such scanners are turned into light stages. However, without such expensive equipment, shape reconstruction often turns out to be erroneous and incomplete owing to occlusions, limited viewpoints, and measurement imperfections. 

Previous work has investigated many different representations of objects to deal with those problems, such as volumetric occupancy grids~\cite{3dshapenets}, distance fields~\cite{deepsdf}, and more recently, point cloud-based representations~\cite{pointnet,pointnet++}. The latter are preferred in terms of efficiency and capacity of maintaining fine-grained details. Yet, each point in a point cloud needs to be represented by a 3D coordinate, and the large number of points generated in reality can strain computational resources, bandwidth, and storage. 

To this end, we propose a novel structured point cloud representation inspired by spherical multi-view 3D object scanners \cite{3dscanner,multicamera}. Sparse point clouds are generated by a simulated configuration of multiple views evenly distributed on a sphere around the object where each view casts a preset bundle of rays from its center towards the object. Due to its similarity to stage lighting instruments, we name it as \textit{Spotlights}. It represents point clouds in a more compact way since it only needs to store a 1D array of scalars, i.e. the depths along the rays, and thus serves as a basis for efficient point-based networks. Moreover, Spotlights is a structured representation that produces ordered point clouds. This order-preserving property is of great benefit to spatial-temporal estimation problems such as Simultaneous Localization And Mapping (SLAM)~\cite{nodeslam} and 3D object tracking~\cite{li2018stereo}.

\begin{figure}
    \centering
    \includegraphics[scale=0.3]{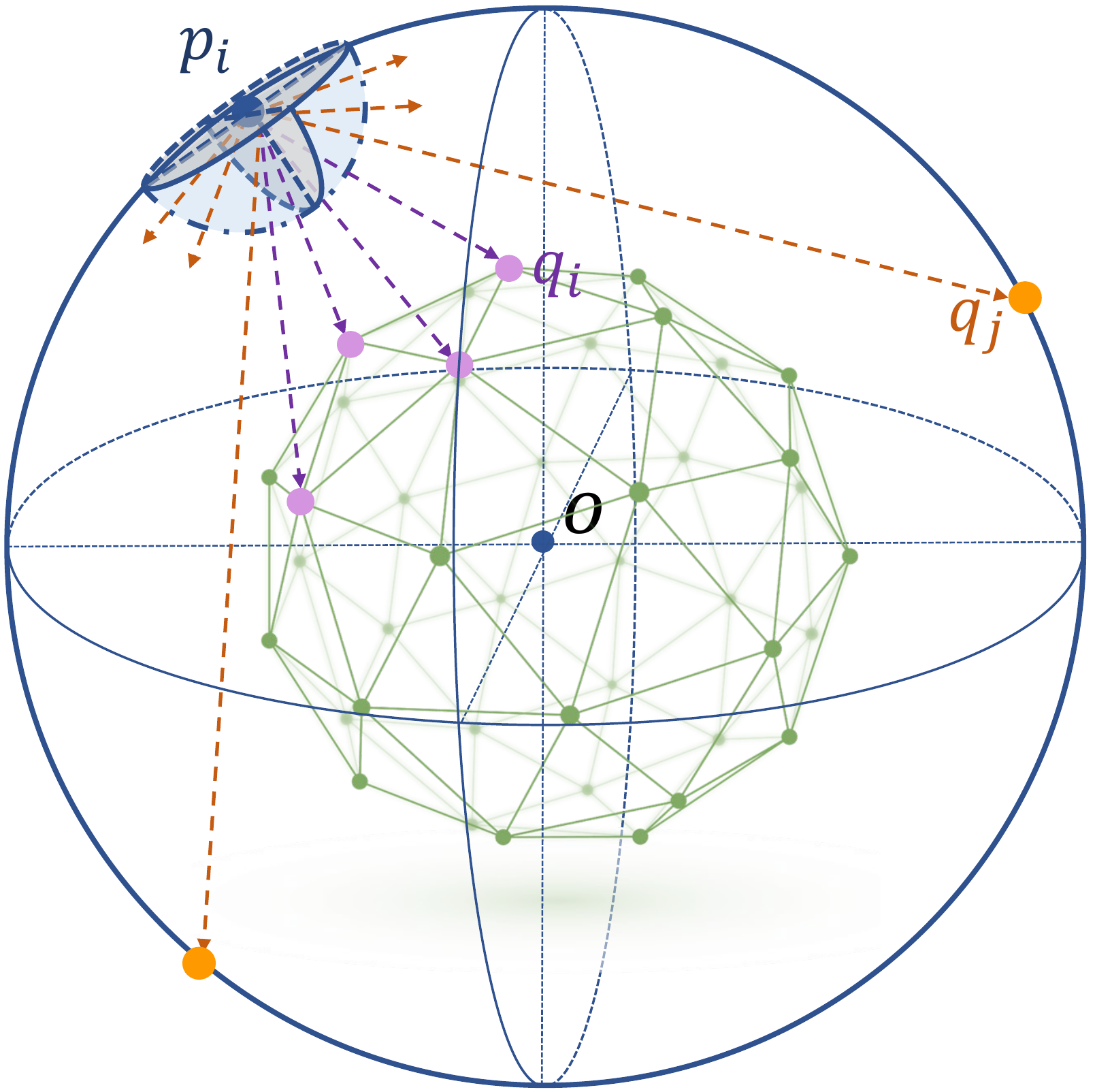}
    \caption{Illustration of Spotlights with only one view. Dashed arrows indicate the ray bundle cast from $p_i$. Some rays intersect with the bounded object (purple) while others spread in void space (orange).}
    \label{fig:3d_model}
\end{figure}

To demonstrate the superiority of Spotlights, we focus on the 3D shape completion task which aims to predict a complete point cloud with respect to a partial observation. Previous efforts are devoted to inferring fine-grained details \cite{dai2017shape,pfnet,pcn} and designing delicate point-based decoders \cite{foldingnet}\cite{topnet}. In this work, we present a lightweight neural network for fast point cloud prediction based on our proposed shape representation. It achieves competitive results in accuracy without substantially sabotaging completeness. Furthermore, our completed point clouds can easily find correspondences across temporal states, therefore, are useful for some subsequent tasks. Our main contributions are as follows:

\begin{itemize}
    \item[$\bullet$] We introduce a novel point representation denoted \textit{Spotlights}. It is a unified and compact representation. The effectiveness of the proposed representation is shown by the theoretical analysis of local point density.
    \item[$\bullet$] We apply our proposed representation to the task of 3D shape completion and achieve remarkable results on par with existing supervised state-of-the-art approaches on both synthetic and real data. Notably, our network consumes less memory and is fast at inference time.
    \item[$\bullet$] We build a new synthetic dataset for 3D shape completion, comprising 128000 training samples and 30000 testing samples over 79 unique car models (i.e. 2000 different observations per object). The ground truth point clouds are extracted by our Spotlights.
\end{itemize}

\section{Related Work}
Nowadays, shape prediction is mostly studied in the context of shape completion and reconstruction from 3D data (e.g. depth images or point clouds). Though this problem can be solved using lower-level geometric principles (e.g. volumetric diffusion~\cite{davis02}, symmetry~\cite{thrun05}) or shape retrieval methods~\cite{li15,pauly05}, the focus of our work lies on learning-based 3D point cloud completion.

In recent years, the community has demonstrated the feasibility of learning highly efficient representations for geometric problems by relying directly on 3D point clouds~\cite{pointnet,pointnet++,shin18}. Fan et al.~\cite{fan2017} and Achlioptas et al.~\cite{achlioptas2018} introduce auto-encoder CNNs for direct point cloud completion, showing that point clouds encode only points on the surface, naturally permit direct manipulation and put fewer restrictions on the topology or fine-grainedness of a shape. Yuan et al.~\cite{pcn} extend the idea by permutation invariant and convolution-free feature extraction, as well as a coarse-to-fine feature generation process. Xie et al.~\cite{xie20} internally revert to a volumetric representation by adding differentiable layers to transform back and forth between point clouds and grids. Huang et al.~\cite{pfnet} furthermore propose to limit the prediction to the missing part of a point cloud, thus preserving details in the original input measurement. Wen et al.~\cite{wen21} mimic an earth mover and complete point clouds by moving points along the surface.

While direct 3D point-based representations have lead to significant advances in terms of efficiency, they suffer from disorder and the inability to focus on inherent local structures. Yang et al.~\cite{foldingnet} propose \textit{FoldingNet}, a deep auto-encoder that generates ordered point clouds by a folding-based decoder that deforms a regular 2D grid of points onto the underlying 3D object surface of a point cloud. Although they claim that their representation can learn ``cuts'' on the 2D grid and represent surfaces that are topologically different from a single plane, they at the same time admit that the representation fails on more complex geometries. Li et al.~\cite{spgan} propose to use GAN to generate realistic shapes from unit spheres, establishing dense correspondences between spheres and generated shapes. Wen et al.~\cite{wen20} propose an extension with hierarchical folding by adding skip-attention connections. Groueix et al.~\cite{atlasnet} introduce \textit{AtlasNet}. Similar to the structure of FoldingNet, this work deforms multiple 2D grids. Nonetheless, the modelling as a finite collection of manifolds still imposes constraints on the overall topology of an object's shape. Tchapmi et al.~\cite{topnet} propose \textit{TopNet}. Their decoder generates structured point clouds using a hierarchical rooted tree structure. The root node embeds the entire point cloud, child nodes subsets of the points, and leaf nodes individual points. Rather than sampling a finite set of manifolds, this architecture generates structured point clouds without any assumptions on the topology of an object's shape. 

\section{Method}
In this section, we first review relevant background knowledge on Fibonacci spheres and solid angles (see Section \ref{Sec:bg}). Then, we describe the detailed construction of our newly proposed Spotlights representation (see Section \ref{Sec:model}), and further prove its theoretical validity (see Section \ref{Sec:math}).

\subsection{Background} \label{Sec:bg}
\subsubsection{Spherical Fibonacci Point Set}
One of the most common approximation algorithms to generate uniformly distributed points on a unit sphere $\mathcal{S}^2 \subset \mathbb{R}^3$ is mapping a 2D Fibonacci lattice onto the spherical surface. The Fibonacci lattice is defined as a unit square $[0, 1)^2$ with an arbitrary number of $n$ points evenly distributed inside it. The 2D point with index $i$ is expressed as
\begin{equation*}
    \mathbf{p}_i = (x_i, y_i) = \left(\left[\frac{i}{\Phi}\right], \frac{i}{n}\right), 0 \leq i < n.
\end{equation*}
Note that $\left[\cdot\right]$ takes the fractional part of the argument and $\Phi = \frac{1 + \sqrt{5}}{2}$ is the golden ratio. The 2D points are then mapped to 3D points on the unit sphere using cylindrical equal-area projection. The 3D points are easily parameterized using the spherical coordinates
\begin{equation*}
    \mathbf{P}_i = (\varphi_i, \theta_i) = (2\pi x_i, \text{arccos}(1-2y_i)), \varphi_i \in [0, 2\pi], \theta_i \in [0, \pi].
\end{equation*}
We refer the readers to \cite{sphericalFibonacci} for more details.

\subsubsection{Solid Angle} Similar to the planar angle in 2D, the solid angle for a spherical cap is a 3D angle subtended by the surface at its center. Suppose the area of the cap is $A$ and the radius is $r$, then the solid angle is given as $\Omega = \frac{A}{r^2}$. Thus, the solid angle of a sphere is $4\pi$. Extended to arbitrary surfaces, the solid angle of a differential area is 
\begin{equation*}
    d\Omega = \frac{dA}{r^2} = \frac{r\text{sin}\theta d\varphi \cdot rd\theta}{r^2} = \text{sin}\theta d\theta d\varphi,
\end{equation*}
where $\theta \in [0, \pi]$ and $\varphi \in [0, 2\pi]$. By calculating the surface integral, we can obtain
\begin{equation*}
    \Omega = \iint \text{sin}\theta d\theta d\varphi.
\end{equation*}

\subsection{Spotlights Model Construction} \label{Sec:model}
In this subsection, we show how to construct the Spotlights model (see Fig. \ref{fig:3d_model} for an illustration). There are three consecutive steps described in the following.
\subsubsection{Bounding Sphere Extraction}
To find the bounding sphere of an object of interest, we first compute its bounding box by determining the difference between the extrema of the surface point coordinates along each dimension. Practically, we can also easily get the bounding box parameters (i.e. center, size and orientation) for real-world objects using state-of-the-art 3D object detection methods \cite{hough,imvotenet,frustum,pointpillars,voxelrcnn}. The bounding sphere is then given as the circumscribed sphere of the bounding box. The radius of the sphere is used as a scaling factor to normalize the real-world object into a unit sphere.

\subsubsection{Ray Bundle Organization}
We leverage the Fibonacci lattice mapping method presented in Section \ref{Sec:bg} to evenly sample the unit sphere obtained from step 1 and place a bundle of virtual viewing rays at each position to mimic the behavior of a camera. The motivation of using a sphere instead of a cuboid arrangement of views is given by the fact that a spherical arrangement has more diversity in the distribution of principal viewing directions, thereby alleviating self-occlusion in the data. Also, spherical multi-view configurations are commonly adopted for 2.5D multi-surface rendering~\cite{pixels,lfdescriptor}.

To define structured ray bundles emitted from each viewpoint, we can imagine placing a smaller sphere at each primary sampling point on the unit sphere. The intersection plane between the unit sphere and each small sphere forms an internal spherical cap, on which we again use Fibonacci sampling to define a fixed arrangement of sampling points. Thus, the rays originate from the primary sampling points and traverse the secondary sampling points on the spherical cap. It is easy to adjust the opening angle of the ray bundle. Suppose the radii of the outer sphere and the small sphere are $R$ and $r$, respectively. The maximum polar angle of the spherical cap is then given by $\omega = \text{arccos}\frac{r}{2R}$. The opening angle $\omega$ as well as the number of rays serve as design hyper-parameters. We refer to Section~\ref{Sec:ablation} for a detailed discussion.

\subsubsection{Ray Casting}
We finally cast the rays onto the object and retrieve the distance between the view center and the first intersection point with the object's surface. For rays that miss the surface, we manually set the distances to a sentinel value of zero. With known origins and viewing directions, the 3D point cloud can be easily recovered from the depths along the rays. Compared to 3D point coordinates, 1D distances are more efficient to store and generate. To validate this property, we apply our representation to the 3D point cloud completion task. While obtaining competitive results in terms of accuracy, our network is fast at inference time and has fewer parameters than previous methods (see Section \ref{Sec:synthetic}). Besides, the fixed ray arrangement enables the reproduction of ordered point clouds, thus providing one-to-one correspondences across temporal states.

\subsection{Analysis of Spotlights} \label{Sec:math}

\begin{figure}
    \centering
    \includegraphics[scale=0.35]{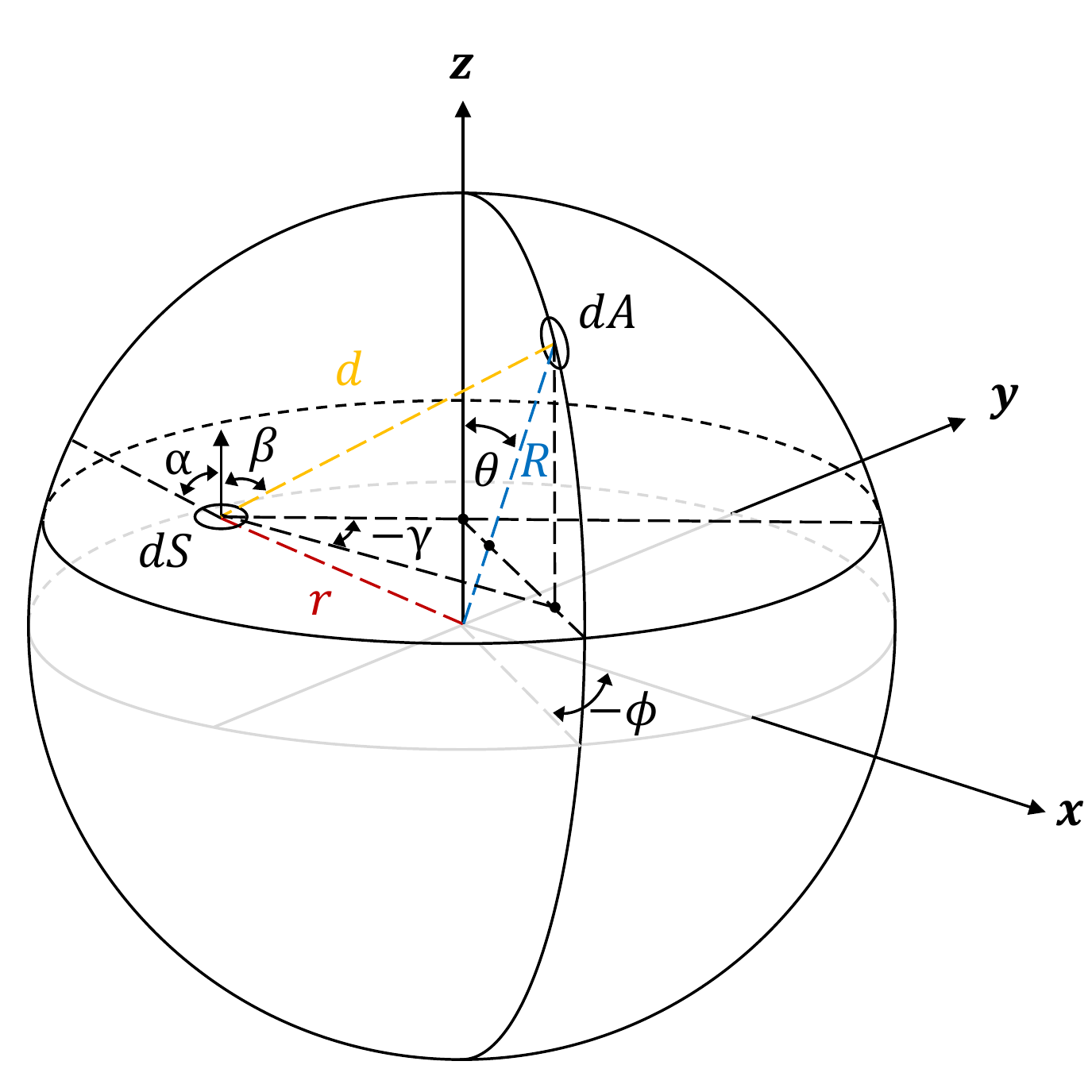}
    \caption{Mathematical model of Spotlights.}\label{Fig:spotlight_model}
\end{figure}

The quality of the sampling is measured from two aspects, namely the uniformity of the ray density for surfels at different positions and orientations within the bounding sphere when 1) there is no occlusion, or 2) there is occlusion. The first measure analyses how much the sampling density of the object surface depends on its location. The second measure provides a means to analyse the impact of occlusion that typically occurs in practical scenarios. To simplify the theoretical analysis, we make the following assumptions:
\begin{enumerate}
    \item The object is bounded by a unit sphere, that is, $R=1$.
    \item An arbitrary surfel $dS$ on the object can be parameterized by the distance of the surfel to the center of the sphere $r$ and the orientation of the surfel $\alpha$, where $\alpha$ is defined as the angle between the surface normal and the vector connecting the centers of the sphere and the surfel (see Fig. \ref{Fig:spotlight_model}). The two parameters are uniformly distributed in their respective ranges, i.e.
    $r\sim U(0, 1)$ and $\alpha \sim U(0, \pi).$
    \item There are $N$ ray sources uniformly distributed on the sphere and each of them emits $M$ rays within a certain solid angle $\Omega$ oriented towards the center of the sphere. For convenience, we set $\Omega = 2\pi$. Note that we also assume $N, M \gg 1$. The source density $\rho_s$ and the ray density $\rho_r$ are then given by
\begin{align*}
    \rho_s = \frac{N}{4\pi R^2} = \frac{N}{4\pi}, \quad\text{and} \quad \rho_r = \frac{M}{2\pi}.
\end{align*}
\end{enumerate}

In order to calculate the ray density of $dS$, we take another surfel $dA$ on the sphere, calculate the ray density of $dS$ caused by $dA$, and then do the integration over $dA$. The area of surfel $dA$ with polar angle $\theta$ and azimuth angle $\phi$ is $dA = \sin \theta d\theta d\phi$. Using trigonometry function, we can calculate the distance between $dS$ and $dA$:
$$d^2 = 1 + r^2 - 2r\left(\cos \alpha \cos \theta - \sin \alpha \sin \theta \cos \phi \right).$$

The rays that can be received by the surfel $dS$ from $dA$ are the solid angle occupied by the view of $dS$ at $dA$ multiplied by the ray density $\rho_r$. Then, the ray density $\rho$ of surfel $dS$ is the integration over $dA$, i.e.
\begin{align*}
    \rho(r, \alpha) &=\frac{1}{dS} \int dA \cdot \rho_s \cdot \rho_r \cdot \frac{dS \cos \beta}{d^2} \\
    &= \frac{NM}{8\pi^2}\int d\theta \int d\phi \frac{\sin \theta \left(\cos \theta - r\cos \alpha \right)}{\left(r^2 - 2r(\cos \alpha \cos \theta - \sin \alpha \sin \theta \cos \phi ) + 1\right)^{3/2}}.
\end{align*}

\begin{figure}
    \centering
    \includegraphics[scale=0.25]{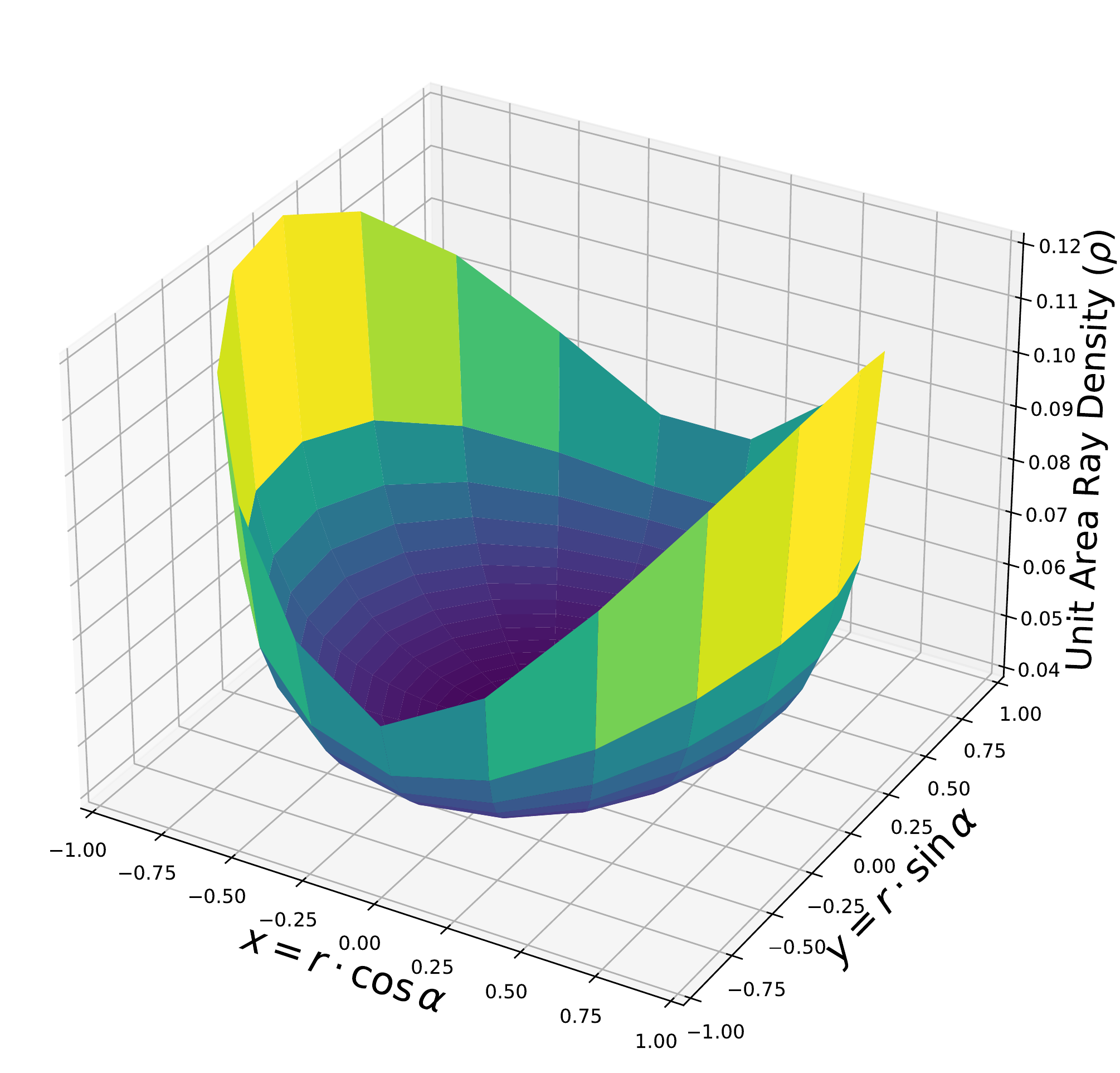}
     \caption{Ray density of surfel $dS$.}\label{Fig:spotlight_density}
\end{figure}

The ray density $\rho$ as a function of $r$ and $\alpha$ is shown in Fig. \ref{Fig:spotlight_density}. We can observe that $\rho$ has no divergence and is a fairly smooth function of $r$ and $\alpha$ when the surfel $dS$ is not too close to the sphere surface. Quantitatively, when $0\leq r\leq 0.8$, $\rho$ takes values between $0.039$ and $0.057$. The proposed sampling strategy therefore ensures that the object surfaces at arbitrary positions and orientations in the sphere have comparable sampling density.

Next, we consider the case where certain directions seen from $dS$ are occluded, therefore no rays can be received from the related sources on the surface of the sphere. The blocked direction is defined by the polar and azimuth angles $(\beta, \gamma)$ (see Fig. \ref{Fig:spotlight_model}). We assume that the blocking is random and uniform, i.e., $\beta \sim U(0, \pi)$ and $\gamma \sim U(0, 2\pi)$. The unit area ray density ($\rho$) as a function of $(r, \alpha, \beta, \gamma)$ is then given by
\begin{align*}
    \rho(r, \alpha, \beta, \gamma) &= \frac{1}{dS} \cdot dA \cdot \rho_s \cdot \rho_r \cdot \frac{dS \cos \beta}{d^2}\\
    &= \frac{NM}{8\pi^2} \frac{\cos{\beta}}{\left(\xi r^2 + 1\right)^{1/2}}, \quad \xi = (\sin \alpha \sin \beta \cos \gamma  - \cos \alpha \cos \beta)^2 - 1.
\end{align*}

As can be observed, $\rho$ is a continuous function of $\beta$ and $\gamma$, which ensures that when part of the directions are occluded, there are still rays that can hit $dS$ from other angles, resulting in better sampling coverage. A simplified version of our point cloud sampling strategy only consists of casting a single ray towards the center of the sphere from each viewpoint. However, it is intuitively clear that this would limit the local inclination of rays sampling $dS$ to a single direction, and thereby lead to an increased likelihood of occlusions. 

Note that in practice the solid angle covered by each bundle of viewing rays should be chosen smaller such that more rays intersect with the object surface, and fewer rays are wasted. We experiment with different solid angles and find a balance between completeness and hit ratio in Section \ref{Sec:ablation}.

\section{Shape Completion with Spotlights}

Next we introduce our 3D shape completion network based on the Spotlights representation, named \textit{Spotlights Array Network (SA-Net)}.

By leveraging the Spotlights model, we reformulate the 3D shape generation into a 1D array regression problem. The ground truth depth values $\mathbf{d}_{gt}$ satisfies that
\begin{equation*}
    \mathbf{P}_{gt} = \mathbf{P}_1 + \mathbf{r} \cdot \mathbf{d}_{gt}, ~\mathrm{and}~ \mathbf{r} = \frac{\mathbf{P}_2 - \mathbf{P}_1}{\Vert \mathbf{P}_2 - \mathbf{P}_1 \Vert}
\end{equation*}
where $\mathbf{P}_{gt}$, $\mathbf{P}_1$ and $\mathbf{P}_2$ are ground truth point cloud, primary sampling points and secondary sampling points, respectively. The ray directions $\mathbf{r}$ are computed using the primary sampling points and secondary sampling points as discussed in Section \ref{Sec:model}. Actually, the Spotlights model is formed by $\mathbf{P}_1$, $\mathbf{P}_2$ and $\mathbf{r}$.

To speed up convergence, we normalize the ground truth depths into the range $[0, 1]$ using a factor $2R$ where $R$ is the radius of the bounding sphere. Missing rays in ground truth are masked with a zero value. Note that the existence of missing rays requires post-processing on the predicted array to filter out the outliers in the recovered point cloud. This is achieved by simply clipping the array using a pre-defined threshold. Here, we empirically set the threshold value to be 0.2.

\begin{figure}
    \centering
    \includegraphics[width=\textwidth]{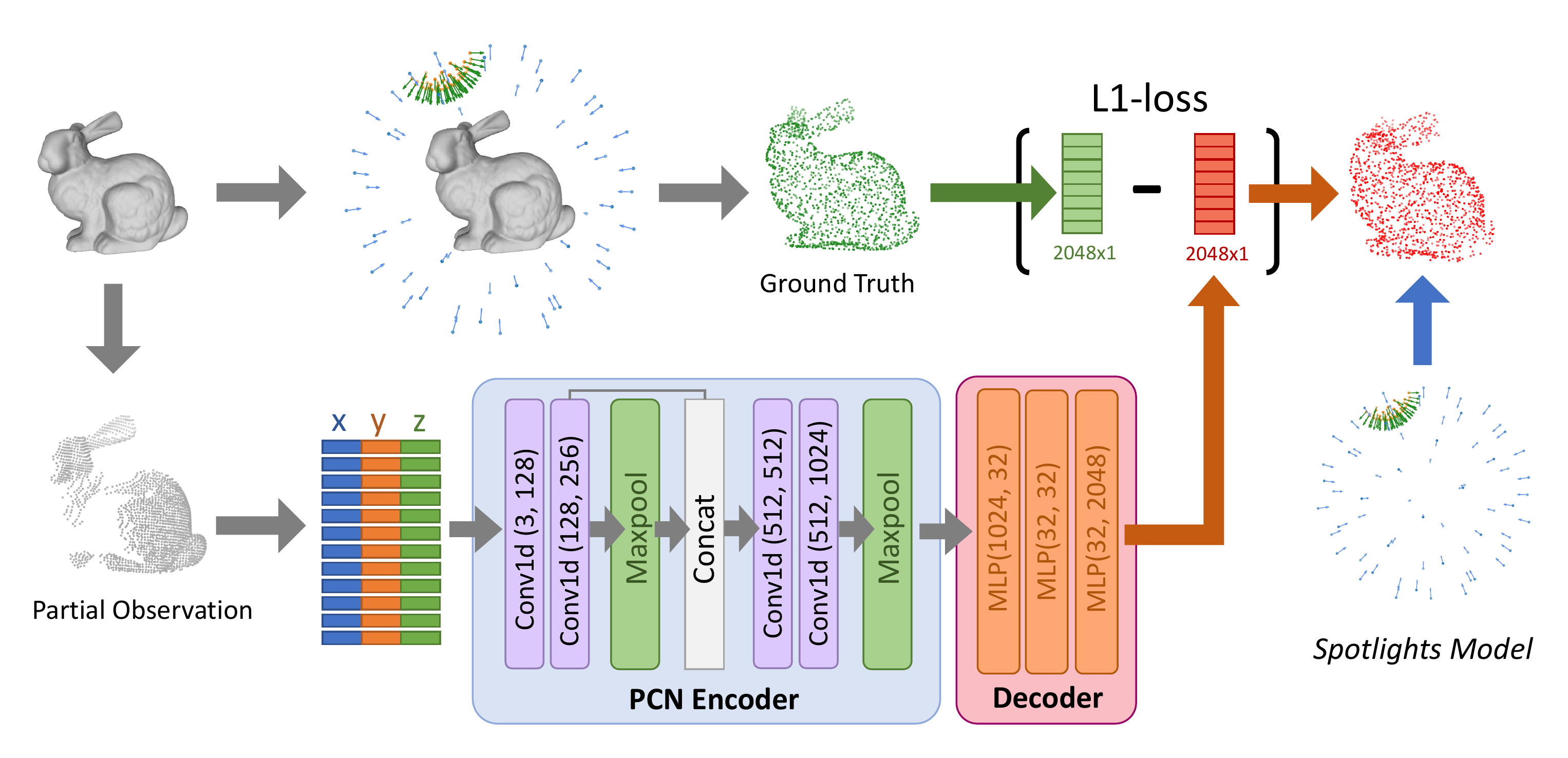}
    \caption{\textbf{Architecture of SA-Net}. The network takes incomplete point cloud observations as input and predicts a 1D array, from which the complete point cloud is recovered by an inverse application of the Spotlights Model. Note that the PCN backbone can be replaced with other point-based encoders.}
    \label{fig:system_overview}
\end{figure}

The pipeline of SA-Net is shown in Fig.~\ref{fig:system_overview}. It consists of a point-based encoder and a decoder purely made up of multi-layer perceptrons (MLPs). Similar to previous work, we feed the partial observation into the encoder to obtain a 1024-dim latent vector. The difference is that we regress an array of scalars instead of 3D coordinates, which significantly reduces the computational cost. We calculate a L1 training loss between the predicted values and ground truth. At inference time, a complete 3D point cloud of the object can be recovered from the predicted 1D scalars using the Spotlights sampling model, that is,
\begin{equation*}
    \mathbf{P}_{pred} = \mathbf{P}_1 + \mathbf{r} \cdot \mathbf{d}_{pred}.
\end{equation*}
In addition, 3D points in $\mathbf{P}_{pred}$ are ordered due to the structure in the Spotlights model.

\section{Experiments}

We first describe how to build a synthetic dataset using Spotlights. Next, we perform 3D shape completion on both the newly made dataset and KITTI-360 \cite{kitti360}, and compare our results against existing state-of-the-art methods. Moreover, we provide ablation studies on ShapeNet \cite{shapenet} to figure out the importance of the hyper-parameters involved in the Spotlights model. 

\subsection{Shape Completion on Synthetic Dataset} \label{Sec:synthetic}
\subsubsection{Data Generation}
\begin{figure}[htbp]
    \centering
    \includegraphics[width=0.7\textwidth]{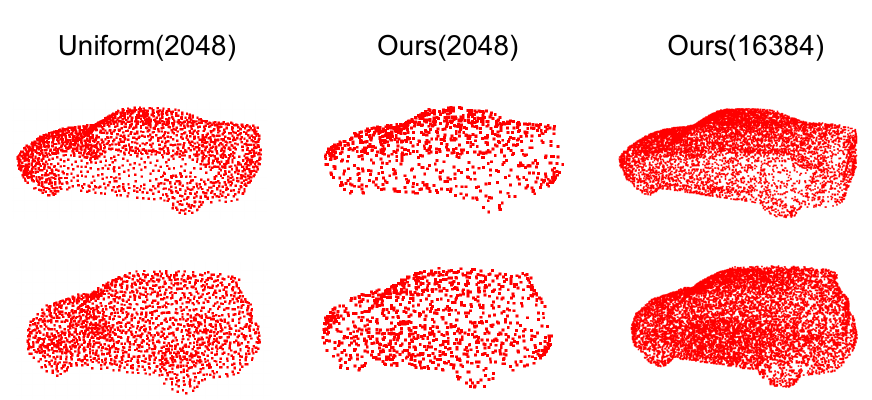}
    \caption{Examples of point clouds generated using different sampling methods. The first column shows the point clouds uniformly sampled from the mesh surface. The second and the third columns show the point clouds sampled by our proposed Spotlights model using different numbers of rays.}
    \label{fig:gt_example}
\end{figure}

We export 79 different vehicle assets of 3 categories from CARLA\cite{carla} and randomly select 64 vehicles for training, 8 vehicles for validation and the rest for testing. We use Blender to simulate a HDL-64 lidar sensor to further generate 2000 partial observations for each object and downsample those point clouds to 256 points. As for the ground truth, we uniformly sample 2048 points from the mesh surface for other methods while using Spotlights to get a 1D array containing 2048 values for SA-Net. Specifically, we evenly sample 32 primary points and 64 secondary points with $60^\circ$ opening angle to construct the Spotlights model. For a fair comparison, we also densely sample 16384 points for each object to calculate the evaluation metrics introduced in the following. Examples of point clouds generated using different sampling methods are shown in Fig.~\ref{fig:gt_example}. Despite the sparsity of the point cloud sampled by Spotlights, it still captures the overall shape of the target object. Besides, increasing the number of rays can notably improve the point density.

\subsubsection{Comparison methods} We compare all methods on the Completion3D benchmark \cite{topnet}, an evaluation platform designed for 3D point cloud completion, and replace the original dataset with our generated synthetic one outlined above. In our setting, TopNet \cite{topnet}, PCN \cite{pcn}, FoldingNet \cite{foldingnet} and AtlasNet \cite{atlasnet} directly predict a complete point cloud containing 2048 points, whereas our SA-Net outputs a 1D array with 2048 values from which we further decode the sparse point cloud. According to the analysis in \cite{pcn}, we use the PCN encoder for all methods.

\subsubsection{Evaluation Metrics} The Chamfer Distance (CD) introduced by \cite{fan2017} is one of the most widely used metrics to measure the discrepancy between a predicted and the ground truth point cloud. Suppose the two point clouds are denoted as $P_1, P_2 \subseteq \mathbb{R}^3$. The CD is given by
\begin{equation}
    d_{CD}(P_1, P_2) = \frac{1}{|P_1|} \sum_{x \in P_1} \mathop{\min}\limits_{y \in P_2} \|x-y\| + \frac{1}{|P_2|} \sum_{y \in P_2} \mathop{\min}\limits_{x \in P_1} \|x-y\|,
    \label{Eq:cd}
\end{equation}
and it computes the average nearest neighbor distance between $P_1$ and $P_2$. Here we use CD in a single direction (i.e. we only employ one of the terms in (\ref{Eq:cd})) to calculate the distance from the prediction to ground truth. As in \cite{acccomp}, this metric indicates the accuracy of each of the predicted points with respect to the ground truth point cloud. To prevent large distances caused by too sparse reference point clouds, we furthermore choose dense ground truth point clouds with 16384 points. In addition, the L1-norm is used to improve robustness against outliers.

\subsubsection{Performance} 
\begin{figure}[htbp]
    \centering
    \includegraphics[width=\textwidth]{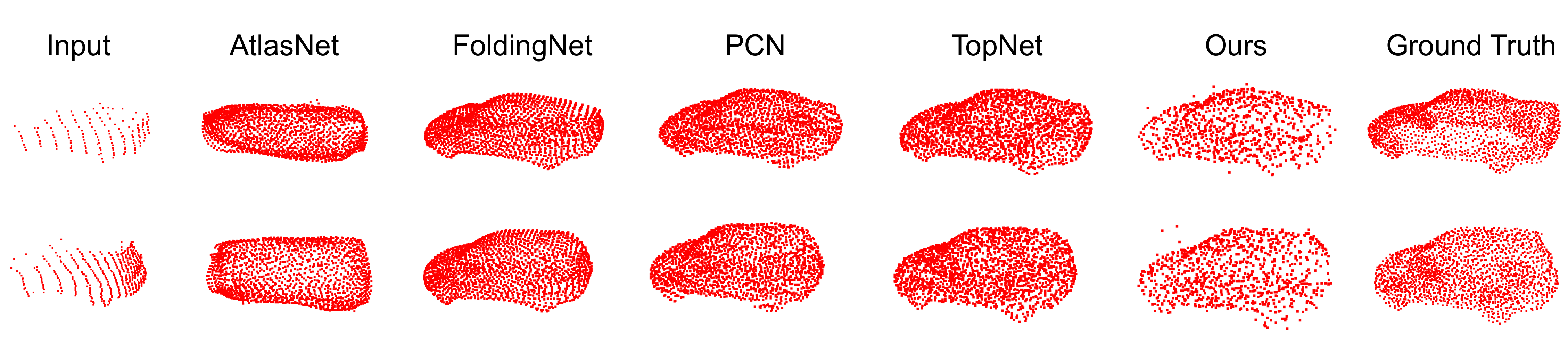}
    \caption{Qualitative Results of shape completion on the synthetic dataset.}
    \label{fig:qualitative_result}
\end{figure}

The quantitative results of SA-Net compared with alternative methods \cite{pcn,foldingnet,topnet,atlasnet} is shown in Table \ref{table:completion_accurarcy}. As can be observed, our SA-Net achieves competitive accuracy, while being faster and using less memory. Note that SA-Net performs better after applying an additional clip operation for statistical outlier removal. The qualitative results are shown in Fig.~\ref{fig:qualitative_result}. The sparsity of our predicted point clouds is caused by missing rays.

\begin{table}[htbp]
\caption{Accuracy and efficiency of comparisons on the synthetic dataset. Accuracy indicates the quality of completed point clouds, which is measured by the single direction CD. Efficiency is evaluated in terms of inference time and the number of network parameters.}
\centering
    \begin{tabular}{c||cccc||c|c}
    \hline
    \multirow{2}{*}{Method} & \multicolumn{4}{c||}{Accuracy} & \multicolumn{2}{c}{Network}\\ \cline{2-7} 
                                    & Hatchback & Sedan & \multicolumn{1}{c||}{SUV} & \multicolumn{1}{c||}{Average} & \multicolumn{1}{c|}{\#Params}  & Runtime\\ \hline\hline
    AtalsNet                        & 0.06586 & 0.07325 & \multicolumn{1}{c||}{0.06051} & 0.06462 & 8.3M & 14.1ms \\
    PCN                             & 0.02528 & 0.02842 & \multicolumn{1}{c||}{0.02221} & 0.02428  & 7.6M & 34.1ms\\
    FoldingNet                         & 0.02437 & 0.02701 & \multicolumn{1}{c||}{0.02166} & 0.02345  & 8.5M & 34.6ms\\
    TopNet                          & 0.02396 & \textbf{0.02631} & \multicolumn{1}{c||}{0.02051} & 0.02252 & 7.5M  & 37.9ms\\ \hline
    Ours (w/o clip)                 & 0.02356 & 0.03095 & \multicolumn{1}{c||}{0.02171} & 0.02442  & \textbf{0.9M} & \textbf{0.80}ms \\
    Ours (w/ clip)                  & \textbf{0.02124} & 0.02775 & \multicolumn{1}{c||}{\textbf{0.01850}} & \textbf{0.02133} & \textbf{0.9M} & 1.05ms \\ \hline
\end{tabular}
\label{table:completion_accurarcy}
\end{table}

\subsection{Shape Completion on Real Dataset}

\begin{figure}[!h]
    \centering
    \includegraphics[width=\textwidth]{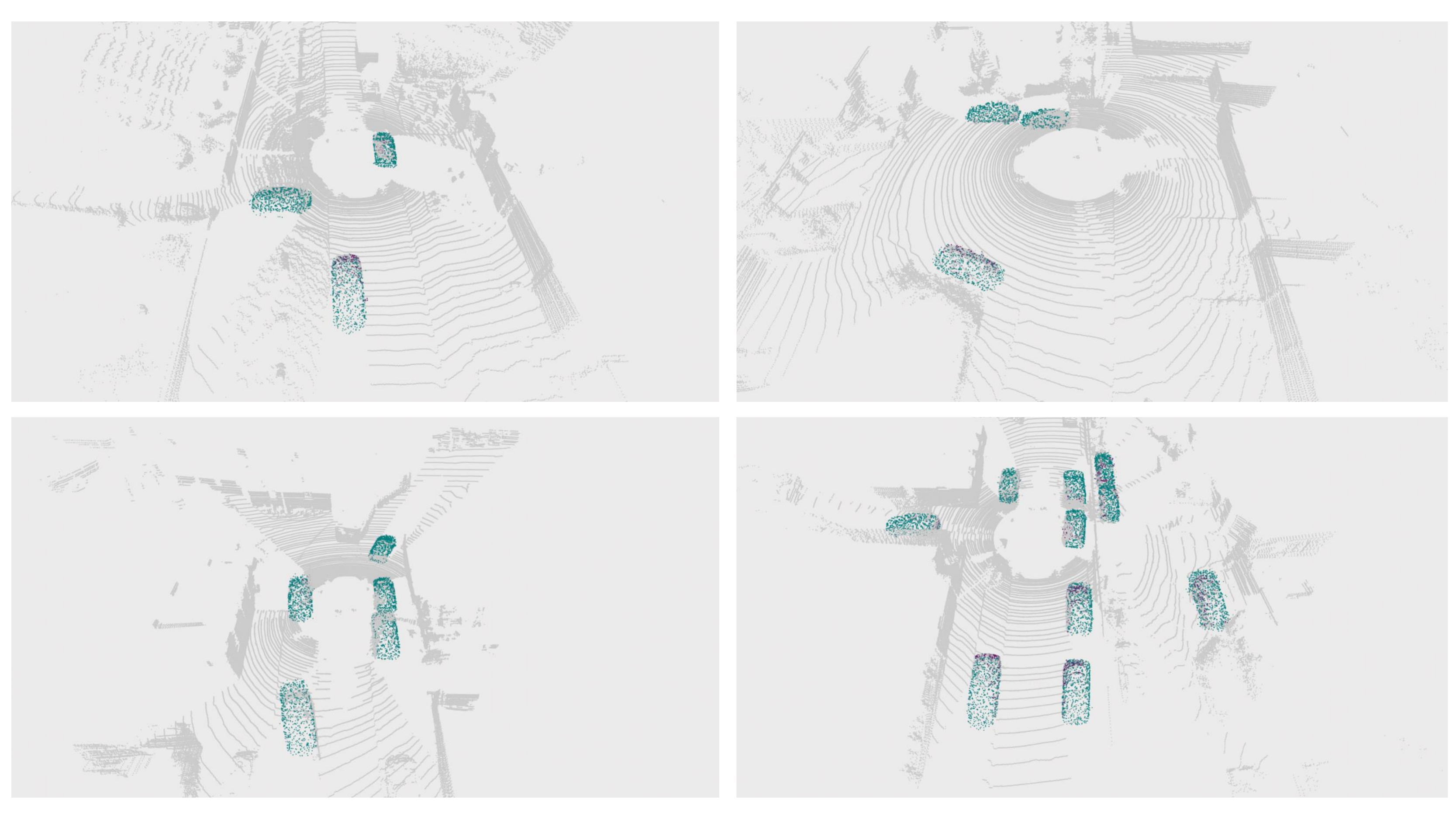}
    \caption{Qualitative results of point cloud completion on real scans in KITTI-360. The gray points are from raw scans. The purple points and the green points are partial inputs and completed results, respectively.}
    \label{fig:real_vis}
\end{figure}

We transfer all models trained on our synthetic dataset to KITTI-360 \cite{kitti360}, a large-scale outdoor dataset, to further evaluate generalization ability. Specifically, we take real-world LiDAR scans and segment out points within the bounding boxes of each frame. This preprocessing generates a total of 4382 partial observations spreading over the classes of \textit{car}, \textit{truck}, \textit{caravan}, and \textit{trailer}. All point clouds are represented in object coordinates and normalized into a unit sphere using the bounding box parameters. Since there is no ground truth point clouds on KITTI-360, we adopt the consistency metric introduced in \cite{pcn}. It averages the CD results over multiple predictions of the same object in different frames. The results are listed in Table \ref{tab:CD}. It is obvious that our SA-Net generates more consistent point clouds, indicating that it can better deal with large variations in the input point clouds. The qualitative results on real scans are shown in Fig.~\ref{fig:real_vis}.

\begin{table}[htbp]
\caption{Consistency of comparisons on KITTI-360. We report the average CD of completed point clouds of the same target to evaluate the consistency of predictions.}
\centering
\label{tab:CD}
\begin{tabular}{c||ccccc}
\hline
\multirow{2}{*}{Model} & \multicolumn{5}{c}{Consistency}                                                                              \\ \cline{2-6} 
                       & Car             & Truck           & Caravan         & \multicolumn{1}{c||}{Trailer}         & Average         \\ \hline\hline
AtalsNet               & 0.0649          & 0.0674          & 0.0651          & \multicolumn{1}{c||}{0.0710}          & 0.0671          \\
PCN                    & 0.0493          & 0.0612          & 0.0644          & \multicolumn{1}{c||}{0.0580}          & 0.0582          \\
FoldingNet                & 0.0514          & 0.0611          & 0.0629          & \multicolumn{1}{c||}{0.0595}          & 0.0587          \\
TopNet                 & 0.0513          & 0.0644          & 0.0648          & \multicolumn{1}{c||}{0.0593}          & 0.0599          \\\hline
Ours                   & \textbf{0.0410} & \textbf{0.0576} & \textbf{0.0613} & \multicolumn{1}{c||}{\textbf{0.0498}} & \textbf{0.0524} \\ \hline
\end{tabular}
\end{table}

\subsection{Ablation Study} \label{Sec:ablation}
In this subsection, we further evaluate the influence of the number of rays, polar angles used in Spotlights. The controlled experiments are conducted on a subset of the ShapeNet dataset \cite{shapenet} containing 6 different categories. Here, we uniformly sample 16384 points on the mesh surface of each object as our ground truth. In addition, the impact of different sampling resolutions is measured through shape completion on our synthetic dataset.

\subsubsection{Completeness of Sampled 3D Point Clouds} Following \cite{acccomp}, we measure the completeness of the 3D point cloud sampled from Spotlights, which indicates how much of the object surface is covered by sample points. The metric again relies on the single-direction CD, however this time iterating through the ground truth point cloud finding the nearest point in the Spotlights representation.

For Spotlights, completeness is closely related to both the number of rays and the polar angle. Hence, we vary the number of rays from 2048 to 8192 and test 3 different polar angles under each setting. Note that the maximum polar angle in our case is about 83 degrees ($r = \frac{1}{4}R$). The average score on each shape category is reported in Table~\ref{Tab:comp}. The results coincide with the intuition that more rays usually produce more points on the surface, whereas polar angles that are too large or too small can lead to degraded point cloud quality since most of the rays will either scatter into the void or concentrate on small local regions.

\begin{table}[!h]
\caption{Completeness of point clouds sampled from Spotlights. We report the average distance from the dense ground truth to the sampled point cloud with varying numbers of rays and polar angles.}
\label{Tab:comp}
\begin{center}
\begin{tabular}{c|c||ccccccc}
\hline
\multirow{2}{*}{\#Rays} & \multirow{2}{*}{Polar angle} & \multicolumn{7}{c}{Completeness}                                                                                                                 \\ \cline{3-9} 
                             &                             & Bathtub         & Bottle          & Cabinet         & Chair           & Lamp            & \multicolumn{1}{c||}{Sofa}            & Average         \\ \hline\hline
\multirow{3}{*}{2048}        & $30^\circ$                          & 0.0474          & 0.0547          & 0.0678          & 0.0444          & 0.0426          & \multicolumn{1}{c||}{0.0619}          & 0.0506          \\
                             & $60^\circ$                         & 0.0349          & 0.0380          & 0.0450          & 0.0339          & 0.0352          & \multicolumn{1}{c||}{0.0447}          & 0.0376          \\
                             & $83^\circ$                         & 0.0382          & 0.0397          & 0.0457          & 0.0389          & 0.0407          & \multicolumn{1}{c||}{0.0469}          & 0.0415          \\ \hline
\multirow{3}{*}{4096}        & $30^\circ$                          & 0.0326          & 0.0367          & 0.0476          & 0.0302          & 0.0298          & \multicolumn{1}{c||}{0.0450}          & 0.0353          \\
                             & $60^\circ$                         & 0.0260          & 0.0278          & 0.0349          & 0.0253          & 0.0263          & \multicolumn{1}{c||}{0.0346}          & 0.0285          \\
                             & $83^\circ$                         & 0.0290          & 0.0297          & 0.0361          & 0.0293          & 0.0312          & \multicolumn{1}{c||}{0.0367}          & 0.0318          \\ \hline
\multirow{3}{*}{8192}        & $30^\circ$                          & 0.0233          & 0.0259          & 0.0345          & 0.0216          & 0.0212          & \multicolumn{1}{c||}{0.0331}          & 0.0255          \\
                             & $60^\circ$                         & \textbf{0.0197} & \textbf{0.0208} & \textbf{0.0279} & \textbf{0.0192} & \textbf{0.0200} & \multicolumn{1}{c||}{\textbf{0.0274}} & \textbf{0.0220} \\
                             & $83^\circ$                         & 0.0222          & 0.0225          & 0.0294          & 0.0223          & 0.0244          & \multicolumn{1}{c||}{0.0295}          & 0.0248          \\ \hline
\end{tabular}
\end{center}
\end{table}

\begin{table}[]
\caption{Hit ratio of rays with different polar angles. It evaluates how many rays actually hit the object.}
\label{Tab:ratio}
\begin{center}
\begin{tabular}{c||ccccccc}
\hline
\multirow{2}{*}{Polar angle} & \multicolumn{7}{c}{Hit ratio (\%)}                                               \\ \cline{2-8} 
                            & Bathtub & Bottle & Cabinet & Chair & Lamp & \multicolumn{1}{c||}{Sofa} & Average \\ \hline\hline
$30^\circ$                          & \textbf{93.8}    & \textbf{81.6}   & \textbf{93.9}    & \textbf{79.1}  & \textbf{47.2} & \multicolumn{1}{c||}{\textbf{85.0}} & \textbf{78.6}    \\
$60^\circ$                         & 63.4    & 45.2   & 67.0    & 49.1  & 31.5 & \multicolumn{1}{c||}{57.1} & 50.9    \\
$83^\circ$                        & 43.2    & 29.9   & 47.7    & 32.6  & 22.6 & \multicolumn{1}{c||}{39.8} & 34.8    \\ \hline
\end{tabular}
\end{center}
\end{table}

\subsubsection{Hit Ratio of Virtual Rays}
In most cases, objects cannot fill the entire space bounded by the outer sphere of Spotlights, which results in wasted rays that have no intersection with the object. To measure the utilization of rays, we compute the hit ratio defined as the number of surface intersecting rays over the total number of casted rays. As shown in Table~\ref{Tab:ratio}, the hit ratio is inversely proportional to the polar angle and much lower for thin objects such as lamps. Although small polar angles waste fewer rays, we strongly suggest not to use those values as completeness also needs to be considered.

\subsubsection{Sampling Resolution}
As discussed in Section \ref{Sec:model}, we sample twice in Spotlights model to better control the ray directions. The total number of rays is then the multiplication of the number of primary and secondary sampling points. To evaluate the performance of Spotlights model with different sampling resolutions, we try several resolution combinations on the task of shape completion and report the accuracy as in Section \ref{Sec:synthetic} in Table \ref{Tab:resolution}. By comparing the first and the second row, we can find that changing the sampling resolutions has no big difference on performance as long as the total number of rays are the same. Also, the third row shows that adding more rays can increase the accuracy, which confirms our intuition.

\begin{table}[!h]
\caption{Accuracy of completed point clouds with different sampling resolutions on the synthetic dataset. For A*B, A and B are the number of primary and secondary sampling points, respectively. Accuracy measures the single direction CD from the prediction to ground truth.}
\label{Tab:resolution}
\begin{center}
    \begin{tabular}{c||cccc}
    \hline
    \multirow{2}{*}{Resolution} & \multicolumn{4}{c}{Accuracy} \\ \cline{2-5} 
                                    & Hatchback & Sedan & \multicolumn{1}{c||}{SUV} & Average \\ \hline\hline
    32*64     & 0.02124 & 0.02775 & \multicolumn{1}{c||}{0.01850} & 0.02133 \\ \hline
    64*32      & 0.02219 & 0.02771 & \multicolumn{1}{c||}{0.01907} & 0.02179 \\ \hline
    256*64      & \textbf{0.02058} & \textbf{0.02700} & \multicolumn{1}{c||}{\textbf{0.01839}} & \textbf{0.02098}  \\ \hline
\end{tabular}
\end{center}
\end{table}

\subsection{Limitations}
Although Spotlights representation is compact compared to 3D coordinate-based point cloud, it still has several limitations. 

First, the ray bundle needs to be carefully organized to reach the balance between efficiency and accuracy. Since the rays spread across the space bounded by the sphere, around 50\% to 60\% of the rays fail to actually hit the object surface, which is a big waste of model capacity and also results in the sparsity of sampled point clouds. 

Second, adjacent rays do not assure intersections in a local surface, which makes it hard to learn depth variation along rays, especially for complex concave objects. To address these issues, in future extensions, we will try a differentiable variant of our Spotlights model to actively learn the orientation of each ray and dynamically adjust it based on the observations. 

\section{Conclusions}

In this paper, we propose a novel compact point cloud representation denoted \textit{Spotlights}, which is useful for applications demanding for high efficiency. In contrast to prior art, the structure in our representation is not imposed in an implicit way using complex network architectures, but in an explicit way through a hand-crafted object surface sampling model, which is highly inspired by the spherical view arrangement in 3D object scanners. Also, no strong assumptions about the enclosed objects are being
made, and a relatively homogeneous sampling distribution is achieved. 

We demonstrate its potential in the context of shape completion where experiments on both synthetic and real dataset show that it can achieve competitive accuracy and consistency while being orders of magnitude faster. The fact that we directly predict depths along rays suggests that our representation has the ability to predict any shape of similar complexity. In future work, we will try to minimize the invalid rays in Spotlights model and figure out a better way to learn the depth variation along the ray.

\bibliographystyle{splncs04}
\bibliography{main}

\end{document}